\documentclass[preprints,article,accept,moreauthors,pdftex]{Definitions/mdpi}  
\pdfoutput=1

\usepackage[utf8]{inputenc}
\usepackage[T1]{fontenc}

\firstpage{1} 
\pubvolume{1}
\issuenum{1}
\articlenumber{0}
\pubyear{2021}
\copyrightyear{2020}
\datereceived{} 
\dateaccepted{} 
\datepublished{} 
\hreflink{https://doi.org/} 

\usepackage{csquotes}
\usepackage[american]{babel}

\usepackage{etoolbox}
\usepackage{physics}
\usepackage{placeins}  

\usepackage{siunitx}
\graphicspath{{figures/}}

\Title{Instance Segmentation of Microscopic Foraminifera}

\TitleCitation{Instance Segmentation of Microscopic Foraminifera}

\Author{
Thomas Haugland Johansen $^{1,*}$\orcidA{},
Steffen Aagaard Sørensen $^{2}$\orcidB{},
Kajsa Møllersen $^{3}$\orcidC{},
Fred Godtliebsen $^{1}$\orcidD{}
}

\AuthorNames{Thomas Haugland Johansen, Steffen Aagaard Sørensen, Kajsa Møllersen and Fred Godtliebsen}

\AuthorCitation{Johansen, T. H.; Sørensen, S. A.; Møllersen, K.; Godtliebsen, F.}

\address{%
$^{1}$ \quad Department of Mathematics and Statistics, UiT The Arctic University of Norway \\
$^{2}$ \quad Department of Geosciences, UiT The Arctic University of Norway \\
$^{3}$ \quad Department of Community Medicine, UiT The Arctic University of Norway
}

\corres{Correspondence: thomas.h.johansen@uit.no}

\abstract{
Foraminifera are single-celled marine organisms that construct shells that remain as fossils in the marine sediments.
Classifying and counting these fossils are important in e.g. paleo-oceanographic and -climatological research.
However, the identification and counting process has been performed manually since the 1800s and is laborious and time-consuming.
In this work, we present a deep learning-based instance segmentation model for classifying, detecting, and segmenting microscopic foraminifera.
Our model is based on the Mask R-CNN architecture, using model weight parameters that have learned on the COCO detection dataset.
We use a fine-tuning approach to adapt the parameters on a novel object detection dataset of more than 7000 microscopic foraminifera and sediment grains.
The model achieves a (COCO-style) average precision of $0.78 \pm 0.00$ on the classification and detection task, and $0.80 \pm 0.00$ on the segmentation task.
When the model is evaluated without challenging sediment grain images, the average precision for both tasks increases to $0.84 \pm 0.00$ and $0.86 \pm 0.00$, respectively.
Prediction results are analyzed both quantitatively and qualitatively and discussed.
Based on our findings we propose several directions for future work, and conclude that our proposed model is an important step towards automating the identification and counting of microscopic foraminifera.
}

\keyword{foraminifera; instance segmentation; object detection; deep learning}

\begin{document}


\section{Introduction}

Foraminifera are microscopic~(typically smaller than 1 mm) single-celled marine organisms~(protists) that during their life cycle construct shells from various materials that readily fossilize in sediments and can be extracted and examined.
Roughly 50~000 species have been recorded of which approximately 9~000 are living today~\cite{Hayward2021}.
Foraminiferal shells are abundant in both modern and ancient sediments.
Establishing the foraminifera faunal composition and distribution in sediments, and measuring the stable isotopic and trace element composition of shell material have been effective techniques for reconstructing past ocean and climate conditions~\cite{Emiliani1955,Hald2007,Katz2010}.
Foraminifera have also proven valuable as bio-indicators for anthropogenically introduced stress to the marine environment~\cite{Suokhrie2017}.
After a sediment core has been retrieved from the seabed, a range of procedures are performed in the laboratory before the foraminiferal specimens can be identified and extracted under the microscope by a geoscientist using a brush or needle.
From each core, several layers are extracted, and each layer is regarded as a sample. 
To establish a statistically robust representation of the fauna, 300-500 specimens are identified and extracted per sample. 
The time-consumption of this task is 2--8 hours/sample, depending on the complexity of the sample and the experience level of the geoscientist.
A typical study consists of 100-200 samples from one or several cores, and the overall time-consumption in just identifying the specimens is vast. 
Recently developed deep learning models show promising results towards automating parts of the identification and extraction process~\cite{Zhong2017,Ge2017,De2017,Mitra2019,Johansen2020}.

\begin{figure}[H]
        \includegraphics[width=\linewidth]{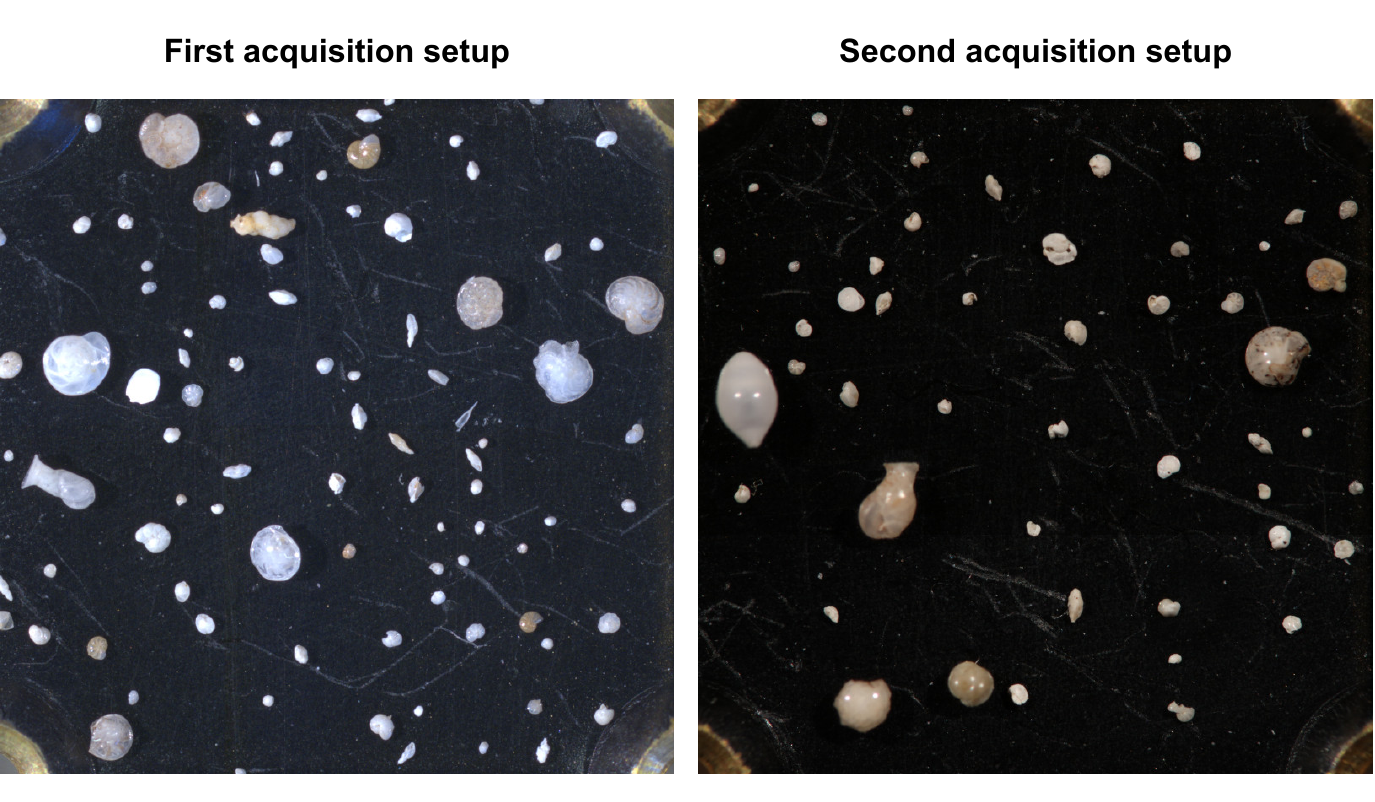}
        \caption{
            Examples of images from the two different image acquisition setups used during three data collection phases.
            \textbf{Left}: Calcareous benthics photographed with the first acquisition setup used during the first data collection phase.
            \textbf{Right}: Calcareous benthics photographed with the second acquisition setup used during the second and third data collection phases.
        }
        \label{fig:examples}
    \end{figure}

Figure~\ref{fig:examples} shows an example of a prepared foraminifera sample --- microscopic objects spread out on a plate and photographed through a microscope. 
Of particular interest is the classification of each object into high-level foraminifera classes, which then serves as input for estimation of the environment in which sediment was produced. 
This task consists of identifying relevant objects, particularly separate sediment from foraminifera, and recognize foraminifera classes based on shapes and structures of each object. 

Object classification in images is one of the great successes of deep learning, and conquer new applications as new methods are developed and high-quality data are made available for training and testing. 
In a deep learning context, a core task of object classification is instance segmentation. 
Not only must the objects be separated from the background, but the objects themselves must be separated from each other, so that adjacent objects are identified, and not treated as one single object. 

Automatic foraminifera identification has great practical potential: the time saved for highly qualified personnel is substantial; it is the overall proportion of foraminifera classes that is the primary interest, and it is therefore robust to the occasional misclassification; and availability of deep learning algorithms that integrates object detection, instance segmentation and object classification.
The lack of publicly available data sets for this particular deep learning application has been an obstacle, but with a curated private data set, soon-to-be published, the stars have finally aligned for an investigation into the potential of applying deep learning to foraminifera classification. 

The manuscript is organized as follows:\\
In Section~\ref{sec:DatasetCuration}, we describe the acquisition and preparation of the dataset, and its final attributes. 
In Section~\ref{sec:InstanceSegmentation}, we give an overview of the Mask R-CNN model applied to foraminifera images. 
In Section~\ref{sec:ExperimentSetup}, we give a detailed description of the experimental setup. 
To present the results, we have chosen to include training behavior (Section~\ref{sec:ModelTraining}), since this is a first attempt for foraminifera application. 
Further, we give a detailed presentation of the performance from various aspects and different thresholds (Section~\ref{sec:ModelPerformance}) for a comprehensive understanding of strengths and weaknesses. 
Section~\ref{sec:Discussion} then emphasizes and discusses the most interesting findings, both in terms of promising performance and in terms of future work (Section~\ref{sec:Future}). 
We round off with a conclusion (Section~\ref{sec:Conclusion}) to condense the discussion into three short statements.

\section{Materials and Methods}

The work presented in this article was performed in two distinct phases;
first a novel object detection dataset of microscopic foraminifera was created, and
then a pretrained Mask~R-CNN model~\cite{He2017} was adapted and fine-tuned on the dataset.

\subsection{Dataset Curation} \label{sec:DatasetCuration}

All presented materials (foraminifera and sediment) were collected from sediment cores retrieved in the Arctic Barents Sea region.
The specimens were picked from sediments influenced by Atlantic, Arctic, polar, and coastal waters representing different ecological environments.
This was done to ensure good representation of the planktic and benthic foraminiferal fauna in the region.
Foraminiferal specimens (planktics, benthics, agglutinated benthics) were picked from the \SIrange{100}{1000}{\micro\metre} size fraction of freeze dried and subsequently wet sieved sediments.
Sediment grains representing a common sediment matrix were also sampled from the \SIrange{100}{1000}{\micro\metre} size range.

The materials were prepared and photographed at three different points in time, with two slightly different image acquisition systems.
During the first two rounds of acquisition, every image contains either pure benthic (agglutinated or calcareous), planktic assemblages, or sediment grains containing no foraminiferal specimens.
In other words, each image contained only specimens belonging to one of four high-level classes; agglutinated benthic, calcareous benthic, planktic, sediment grain.
This approach greatly simplified the task of labeling each individual specimen with the correct class.
In order to better mimic a real-world setting with mixed objects, the third acquisition only contained images where there was a realistic mixture of the four object classes.
To get the necessary level of magnification and detail, four overlapping images were captured from the plates on which the specimens were placed, where each image corresponded to a distinct quadrant of the plate.
The final images were produced by stitching together the mosaic of the four partially overlapping images.


All images from the first acquisition were captured with a 5 megapixel Leica~DFC450 digital camera mounted on a Leica~Z16~APO fully apochromatic zoom system.
The remaining two acquisitions were captured using a 51 megapixel Canon~EOS~5DS~R camera mounted on a Leica~M420 microscope.
The same Leica~CLS~150x (twin goose-neck combination light guide) was used for all acquisitions, but with slightly different settings.
The light power was set to 4 and 4 for the first acquisition, and to 3 and 3 for the remaining two.
No illumination or color correction was performed, in an attempt to mimic a real-world scenario of directly detecting, classifying and segmenting foraminifera placed under a microscope.
Examples of the differences in illumination settings can be seen in Figure~\ref{fig:examples}.

\begin{figure}[H]
        \includegraphics[width=\linewidth]{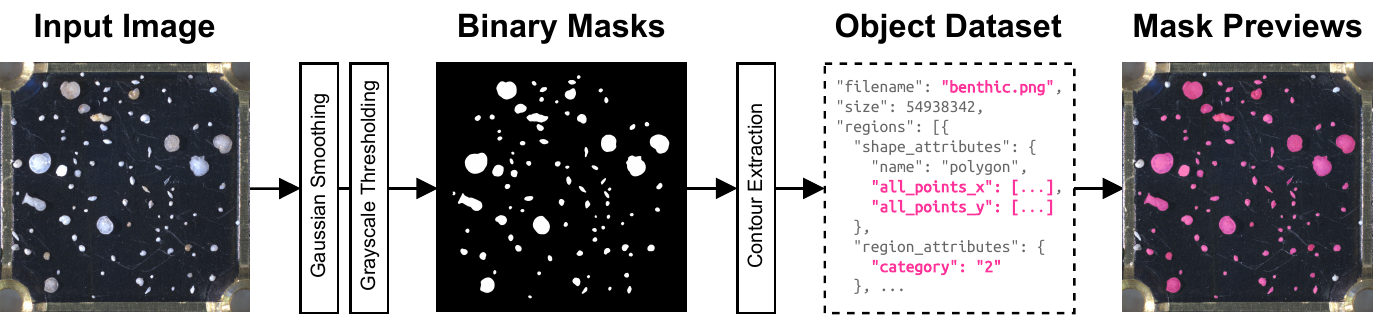}
        \caption{
            High-level summary of the dataset creation pipeline.
        }
        \label{fig:pipeline}
    \end{figure}

To create the ground truth, a simple, yet effective, hand-crafted object detection pipeline~\cite{Johansen2020} was ran on each image, which produced initial segmentation mask candidates.
The pipeline consisted of two steps of Gaussian smoothing, then grayscale thresholding followed by a connected components approach to detect individual specimens.
Some parameters such as the width of Gaussian filter kernel, and threshold levels, were hand-tuned to produce good results for each image in the dataset.
A simple illustration of the preprocessing pipeline can be seen in Figure~\ref{fig:pipeline}. For full details, see \citet{Johansen2020}.

After obtaining the initial segmentation mask dataset, all masks were manually verified and adjusted using the VGG~Image~Annotator~\cite{Dutta2016,Dutta2019} software.
Additionally, approximately 2000 segmentation masks were manually created (using the same software) for objects not detected by the detection pipeline.
The end result is a novel object detection dataset consisting of 104 images containing over 7000 segmented objects.
Full details on the final dataset can be found in Table~\ref{tab:dataset}.

    \begin{specialtable}[H]
        \caption{
            Detailed breakdown of the object dataset, where row holds information for a specific microscope image acquisition phase.
            The first and second phases contain only \enquote{pure} images where every object is of a single class, whereas the third phase images contain only mixtures of several classes.
        }
        \label{tab:dataset}
        \begin{tabular}{@{}lrrrrrr@{}}
        \toprule
                 &     &      & \multicolumn{4}{c}{\textbf{Objects per class}} \\ \cmidrule(l){4-7} 
        \textbf{Phase} & \textbf{Images} & \textbf{Objects} & \textbf{Agglutinated} & \textbf{Benthic} & \textbf{Planktic} & \textbf{Sediment} \\ \midrule
        First    & 48  & 3775 & 172       & 897        & 726       & 1980      \\
        Second   & 41  & 2604 & 583       & 695        & 657       & 669       \\
        Third    & 15  & 633  & 154       & 156        & 155       & 168       \\ \midrule
        Combined & 104 & 7012 & 909       & 1748       & 1538      & 2817      \\ \bottomrule
        \end{tabular}
    \end{specialtable}

\subsection{Instance Segmentation using Deep Learning} \label{sec:InstanceSegmentation}

\begin{figure}[H]
        \includegraphics[width=\linewidth]{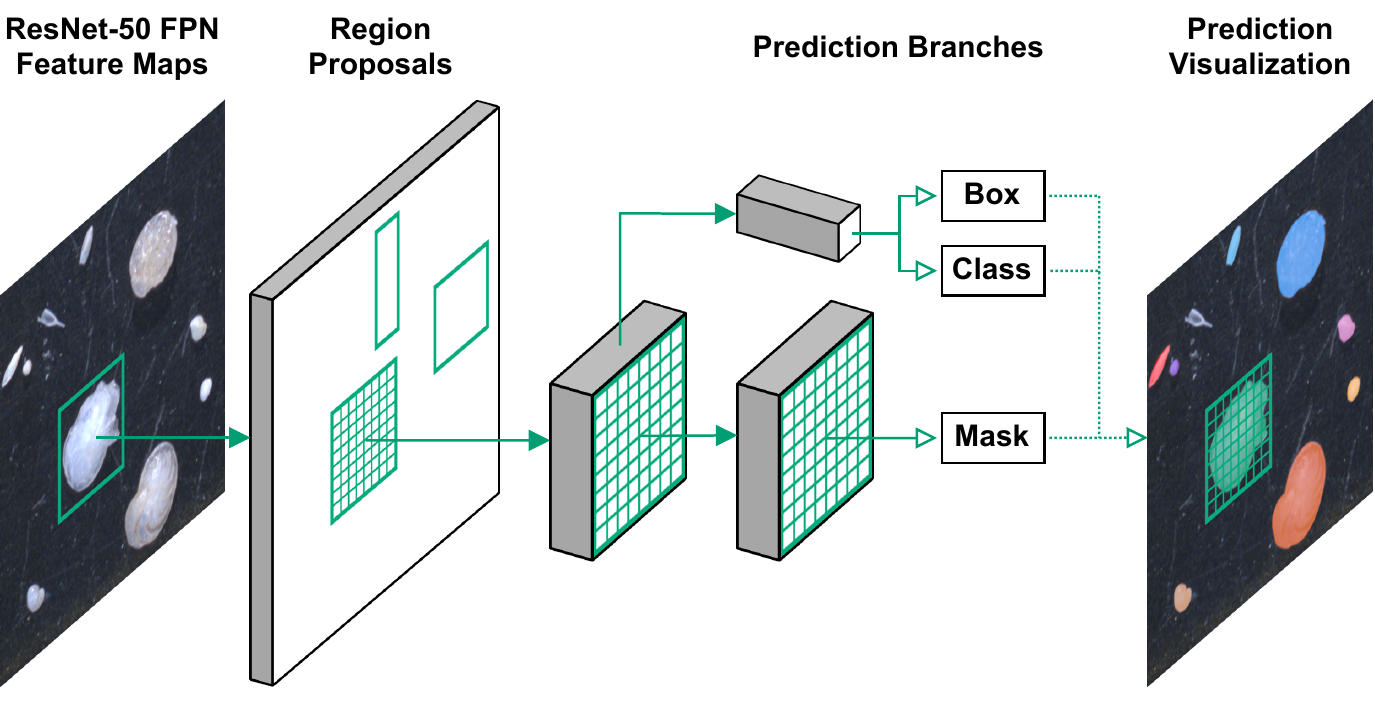}
        \caption{
            Simple, sketch-like depiction of the Mask R-CNN model architecture.
          }
        \label{fig:model}
    \end{figure}

Mask~R-CNN~\cite{He2017} is a proposal-based deep learning framework for instance segmentation, and is an extension to Fast/Faster~R-CNN~\cite{Girshick2015,Ren2017}.
In the Fast/Faster~R-CNN framework the model has two output branches, one that performs bounding box regression and another that performs classification.
The input to these two branches are pooled regions of interest~(RoIs) produced from features extracted by a convolutional neural network~(CNN) backbone.
This is extended in Mask~R-CNN by adding an extra (decoupled) output branch, which predicts segmentation masks on each RoI.
Figure~\ref{fig:model} shows a simple, high-level representation of the Mask~R-CNN model architecture.
Several alternatives to Mask~R-CNN exist, such as PANet~\cite{Liu2018}, TensorMask~\cite{Chen2019}, CenterMask~\cite{Wang2020}, and SOLOv2~\cite{XWang2020}.
We chose to use the Mask~R-CNN framework for two key reasons:
(1) it predicts bounding boxes, class labels, and segmentation masks at the same time in a single forward-pass, and
(2) pretrained model parameters are readily available, removing the need to train the model from scratch.


Due to its flexible architecture, there are numerous ways to design the feature extraction backbone of a Mask~R-CNN model.
We chose a model design based on a ResNet-50~\cite{He2015} Feature~Pyramid~Network~(FPN)~\cite{Lin2017} backbone for feature extraction and RoI proposals.
To avoid having to train the model from scratch, we applied model parameters pretrained on the COCO~dataset~\cite{Lin2015}.
The object detection model and all experiment code was implemented using Python 3.8, PyTorch 1.7.1, and torchvision 0.8.2.
The pretrained model weights were downloaded via the torchvision library.

\subsection{Experiment Setup and Training Details} \label{sec:ExperimentSetup}

The original Mask~R-CNN model was trained using 8 GPUs and a batch size of 16 images, with 2 images per GPU.
We did not have access to that kind of compute resources, and were instead limited to a single NVIDIA TITAN Xp GPU, which also meant our training batches only consisted of a single image.
The end result of this was slightly more unstable loss terms and gradients, so we carefully tested many different optimization methods, learning rates, learning rate scheduling, and so on.

The dataset was split (with class-level stratification) into separate training and test sets, using a $2.47:1$ ratio, which produced 74 training images and 30 test images.
The training and test sets remained the same for all experiments.
During training images were randomly augmented, which included horizontal and vertical flipping, brightness, contrast, saturation, hue, and gamma adjustments.
Both the horizontal and vertical flips were applied independently, with a flip probability of $50\%$ for both cases.
Brightness and contrast factors were randomly sampled from $\qty[0.9, 1.1]$, the saturation factor from $\qty[0.99, 1.01]$, and hue from $\qty[-0.01, 0.01]$.
For the random gamma augmentation, the gamma exponent was randomly sampled from $\qty[0.8, 1.2]$.

We ran the initial experiments using the Stochastic Gradient Descent~(SGD) optimization method with Nesterov~momentum~\cite{Sutskever2013} and weight decay.
The learning rates tested were $\qty{\num{e-3}, \num{5e-3}, \num{e-5}}$, and the momentum parameter was set $0.9$.
For weight decay, we tested the values $\qty{0, \num{e-4}, \num{e-5}, \num{5e-5}}$.
In some experiments the learning rate was reduced by a factor of $10$ after either 15 or 25 epochs.
Training was stopped after 50 epochs.
After the initial experiments with SGD we tested the Adam~\cite{Kingma2015} optimization method.
We tested the learning rates $\qty{\num{e-3}, \num{e-4}, \num{e-5}, \num{e-6}}$.
The weight decay parameter values were $\qty{0, \num{e-4}, \num{e-6}}$.
We used the same scheduled learning rate decay as with SGD for the initial experiment, and training was stopped after 50 epochs.

From on our initial experiments with SGD and Adam, we saw that the latter gave more stable loss terms during training and decided to use this method for the subsequent experiments.
However, we experimented with a recent variant of the Adam optimizer with decoupled weight decay, referred to as AdamW~\cite{Loshchilov2019}.
We implemented a slightly adjusted scheduled learning rate decay, with a factor of 10 reduction after both 25 and 45 epochs of training.
Because we used model parameters pretrained on the COCO dataset, we also ran experiments with fine-tuning the backbone model to adapt it to our target domain.
For the fine-tuning experiments we tested when to ``freeze``and ``unfreeze'' the backbone model parameters, i.e. when to fine-tune the backbone, as well as which layers of the backbone to fine-tune.

Based on all probing experiments and the hyperparameter tuning, our final model was trained using AdamW for 50 epochs.
During the first 25 epochs of training the last three ResNet-50 backbone layers were fine-tuned, and then subsequently frozen.
The initial learning rate was set to $\num{e-5}$ and was reduced to $\num{e-6}$ after 25 epochs, and further reduced to $\num{e-7}$ after 45 epochs.
We set the weight decay parameter to $\num{e-4}$.
Using this configuration, we trained the model 10 times using different random number generator states to ensure valid results and to measure the robustness of the model.

\section{Results}

Model performance is evaluated using the standard COCO metrics for object detection and instance segmentation~\cite{COCO2021}.
Specifically, we are using the average~precision~(AP) and average~recall~(AR) metrics averaged over 10 intersection-over-union~(IoU) thresholds\footnotemark and all classes.
\footnotetext{The IoU thresholds range from $0.5$ to $0.95$ in increments of $0.05$.}
We also use the more traditional definition of AP, which is evaluated at a specific IoU, e.g. AP$_{50}$ denotes the AP evaluated with an IoU of 0.5.
Additionally, we present conventional precision-recall curves with different evaluation configurations, e.g. per-class, per-IoU, and so forth.
All presented precision and recall results were produced by evaluating models on the test split of the dataset.

\subsection{Model Training} \label{sec:ModelTraining}

\begin{figure}[H]
        \includegraphics[width=\linewidth]{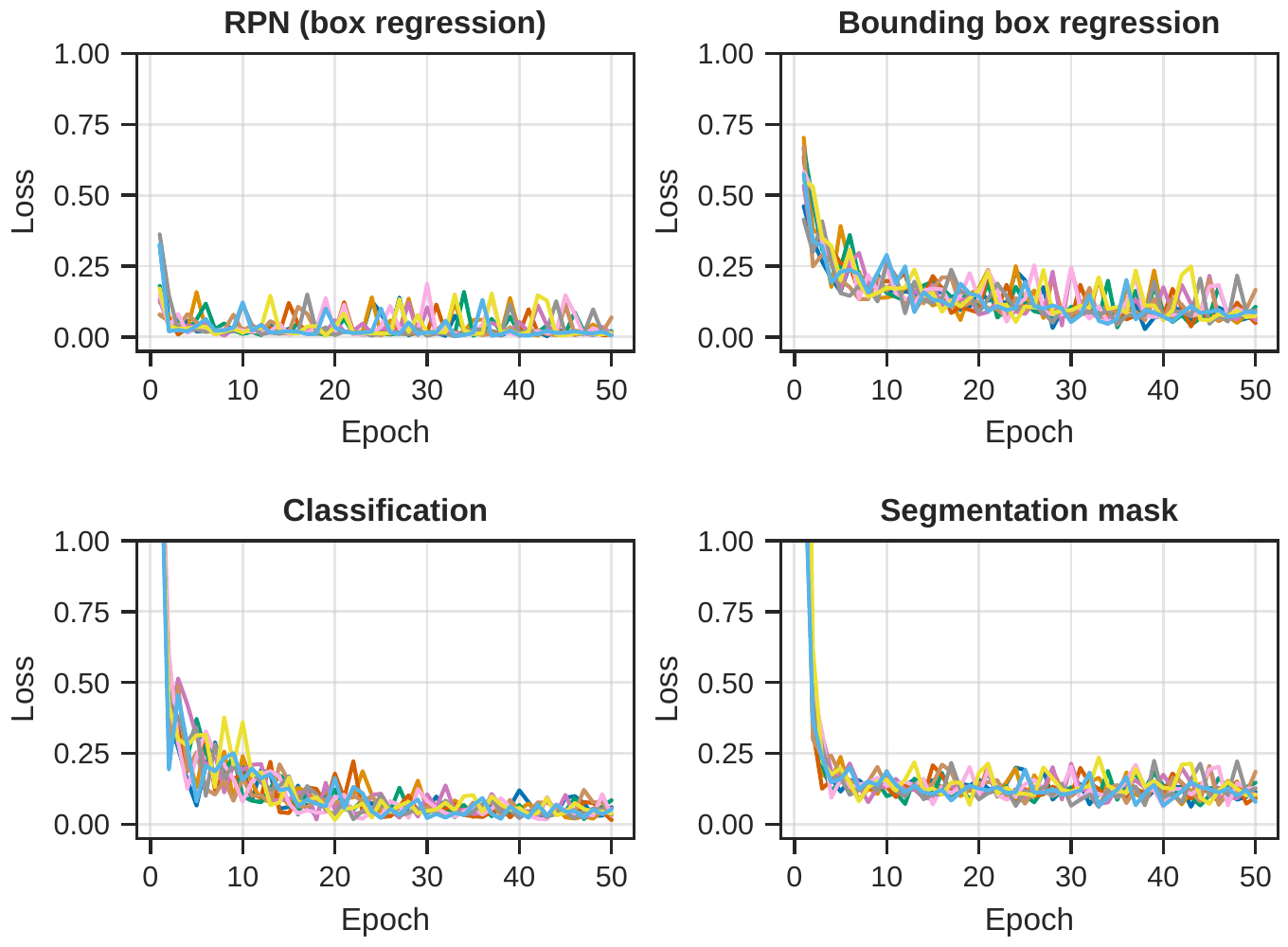}
        \caption{
            Evolution of the individual loss terms for each of the 10 training runs.
            \textbf{Top left}: The loss term for the RPN box regression sub-task.
                Fairly rapid convergence, but we can see the effect of the single-image batches in the curves.
            \textbf{Top right}: Bounding box regression loss for the detection branch.
                The convergence is slower when compared to the RPN loss, and perhaps slightly less stable.
            \textbf{Bottom left}: Object classification loss for the detection branch.
                Similar story can be seen here as with the bounding box regression, which suggests a possible challenge with the detection branch.
            \textbf{Bottom right}: The segmentation mask loss for the segmentation branch.
                Fast convergence, but again we see the effect of the single-image training batches.
        }
        \label{fig:training_loss}
    \end{figure}

\begin{figure}[H]
        \includegraphics[width=\linewidth]{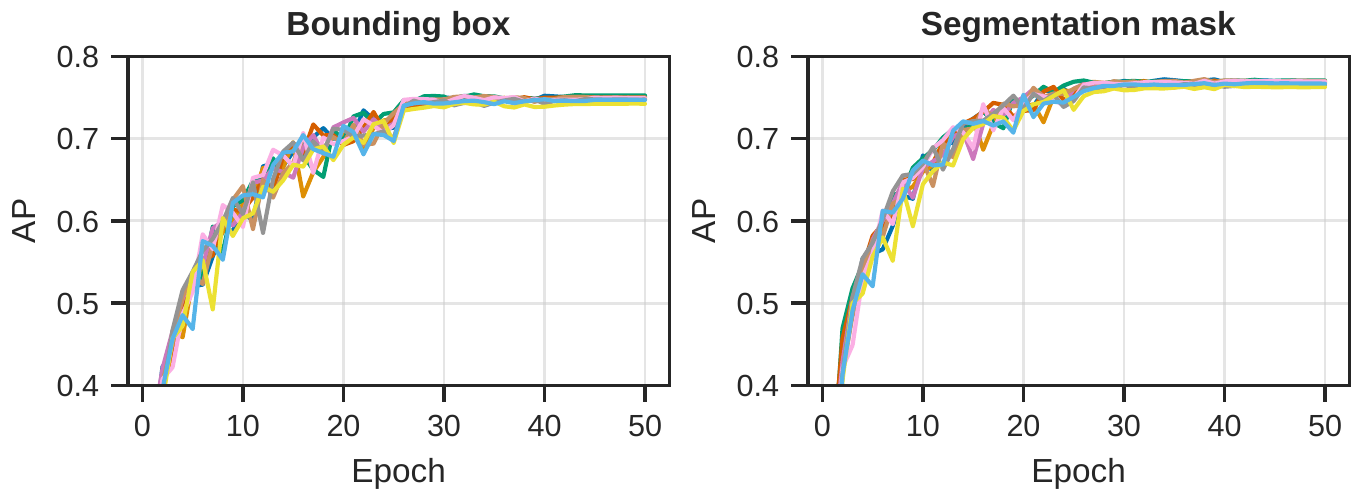}
        \caption{
            Evolution of the AP for both the detection and segmentation task, for each of the 10 runs, per epoch of training.
            Note that these results are based on evaluations using a maximum of 100 detections per image.
            \textbf{Left}: The AP for the detection (bounding box) task. A plateau is reached after about 30 epochs.
            \textbf{Right}: The AP for the segmentation mask task. We observe that the same type of plateau is reached here as with the detection task.
        }
        \label{fig:ap_suite}
    \end{figure}

During training, all training losses were carefully monitored and reported both per-batch and per-epoch.
Four of the key loss terms for the Mask R-CNN model can be seen in Figure~\ref{fig:training_loss}, where each curve represents one of the 10 repeated training runs, with different initial random state.
At the end of every training epoch, we evaluated the model performance in terms of the AP metric for both the detection and segmentation task on the test images.
The per-epoch results for all 10 runs can be seen in Figure~\ref{fig:ap_suite}.

These results indicate that even though we reached some kind of plateau during training, we did not end up overfitting or otherwise hurt the performance on the test dataset.
The AP for both tasks also reach a plateau, which is almost identical for all of the learned model parameters.
This suggests that the training runs reached an upper limit on performance given the dataset, model design, and hyperparameters.

\subsection{Evaluating the Model Performance} \label{sec:ModelPerformance}

After the 10 training runs had concluded, we evaluated each model on the test data using their respective parameters from the final training epoch.
Note that all precision and recall evaluations presented from this point onward are based on a maximum of 256 detections per image\footnote{During training we ran model evaluations with a maximum of 100 detections per image due to lower computational costs and faster evaluation.}.
The mean AP across all of the 10 run was evaluated as $0.78 \pm 0.00$ for the detection task, and $0.80 \pm 0.00$ for the segmentation task.
Table~\ref{tab:average_scores} shows a summary of the AP and AR metrics for both tasks, where each result is the mean and standard deviation of all training runs.
Note that this table shows results averaged over all four classes, and also with the \enquote{sediment} class omitted from each respective evaluation, which will be discussed later.

    \begin{specialtable}[H]
        \caption{
            AP and AR scores for different IoU thresholds, evaluated with all object classes being considered and with the \enquote{sediment} class excluded.
        }
        \label{tab:average_scores}
        \begin{tabular}{@{}l r r r r@{}}
            \toprule
            & \multicolumn{2}{c}{\textbf{All classes}}  & \multicolumn{2}{c}{\textbf{Sans ``sediment'' class}}  \\
            \cmidrule(l){2-5} 
            & \textbf{Bound. box}  & \textbf{Segm. mask}  & \textbf{Bound. box}  & \textbf{Segm. mask}  \\
            \midrule
            $\mathrm{AP}_{50}$ & $0.90 \pm 0.00$  & $0.90 \pm 0.00$  & $0.94 \pm 0.00$  & $0.95 \pm 0.00$  \\
            $\mathrm{AP}_{75}$ & $0.88 \pm 0.00$  & $0.90 \pm 0.00$  & $0.94 \pm 0.00$  & $0.94 \pm 0.00$  \\
            $\mathrm{AP}$      & $0.78 \pm 0.00$  & $0.80 \pm 0.00$  & $0.84 \pm 0.00$  & $0.86 \pm 0.00$  \\
            \midrule
            $\mathrm{AR}$      & $0.83 \pm 0.00$  & $0.84 \pm 0.00$  & $0.89 \pm 0.00$  & $0.90 \pm 0.00$  \\
            \bottomrule
        \end{tabular}
    \end{specialtable}

\begin{figure}[H]
        \includegraphics[width=\linewidth]{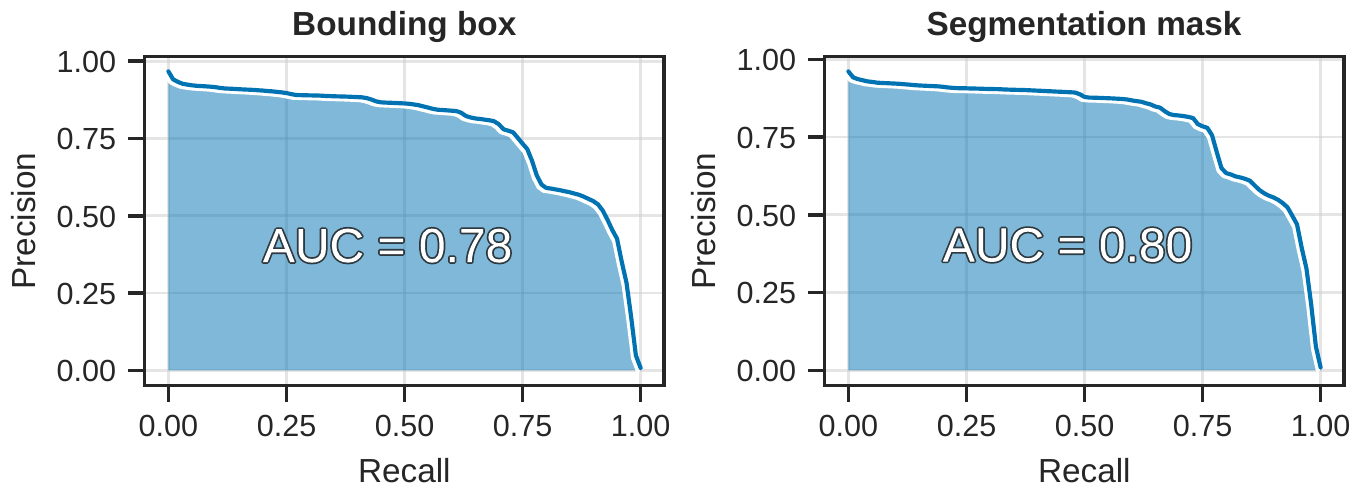}
        \caption{
            The mean average precision-recall curves for the 10 training runs, for both the detection and segmentation tasks.
            The area-under-the-curve~(AUC) shown here is the same as our definition of AP, which can be seen in Table~\ref{tab:average_scores}.
        }
        \label{fig:pr_average}
    \end{figure}

The precision-recall curves computed by averaging over all 10 training runs can be seen in Figure~\ref{fig:pr_average}.
Note the sharp and sudden drop in the curve around the recall threshold of 0.75, for both tasks.

\begin{figure}[H]
        \includegraphics[width=\linewidth]{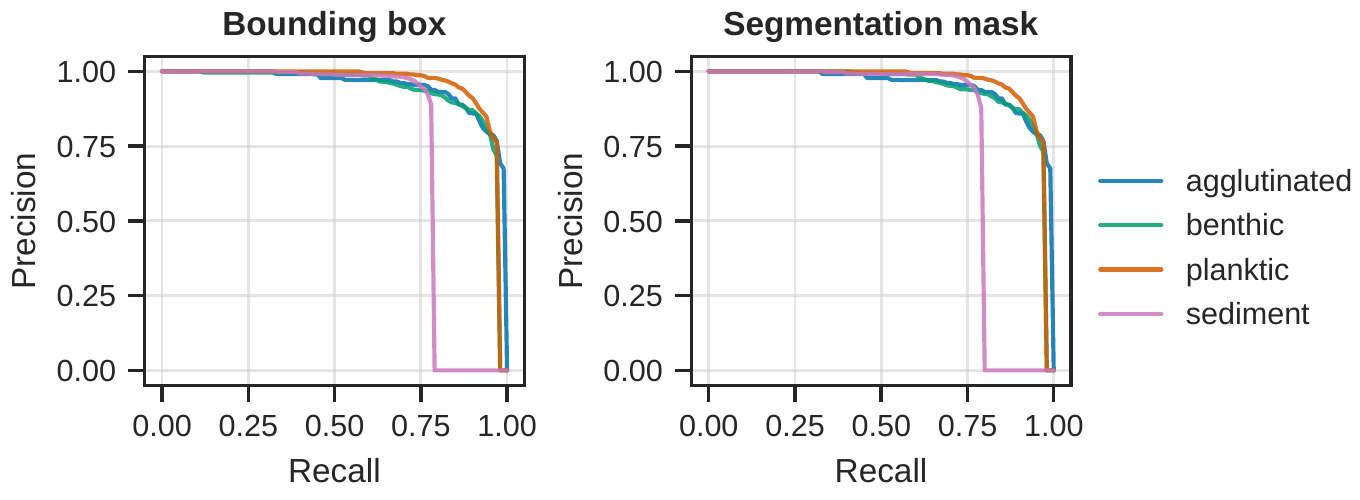}
        \caption{
            Precision-recall curves for each of the four object classes.
            \textbf{Left}: Per-class curves for the detection (bounding box) task.
                Performance is approximately the same for the \enquote{agglutinated}, \enquote{benthic}, and \enquote{planktic} classes,
                but is significantly worse for the \enquote{sediment} class.
            \textbf{Right}: The per-class curves for the segmentation task, which tells the same story as for the detection task.
        }
        \label{fig:pr_classes}
    \end{figure}

In order to investigate the sharp drop in precision and recall, we computed per-class precision and recall; the results can be seen in Figure~\ref{fig:pr_classes}.
From the curves in the figure it is clear that the model is finding the \enquote{sediment} class particularly challenging.
Notice how the precision rapidly goes towards zero slightly after the recall threshold of $0.75$.

\begin{figure}[H]
        \includegraphics[width=\linewidth]{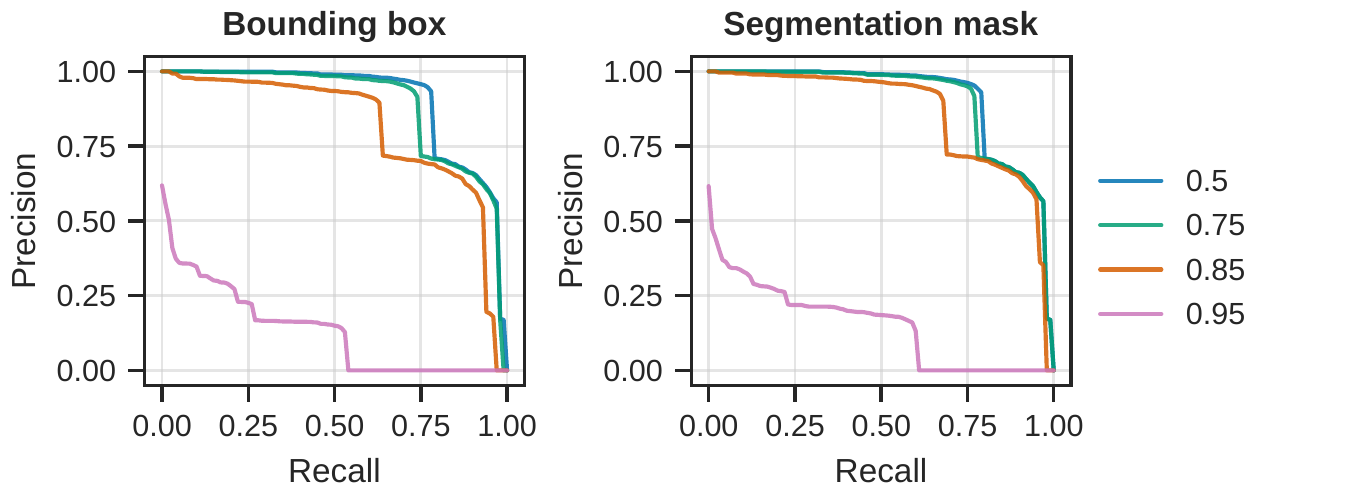}
        \caption{
            Precision-recall curves at different IoU thresholds, where each curve is based on the average for all four object classes.
            \textbf{Left}: PR curves for the detection (bounding box) task.
                There is a sharp drop in precision at the approximate recall thresholds $\qty{0.65, 0.74, 0.8}$, which corresponds to the lower precision of the \enquote{sediment} class.
            \textbf{Right:} The same drop in precision is observer for the predicted masks, which can again be explained by the performance on the \enquote{sediment} class.
        }
        \label{fig:pr_iou}
    \end{figure}

\begin{figure}[H]
        \includegraphics[width=\linewidth]{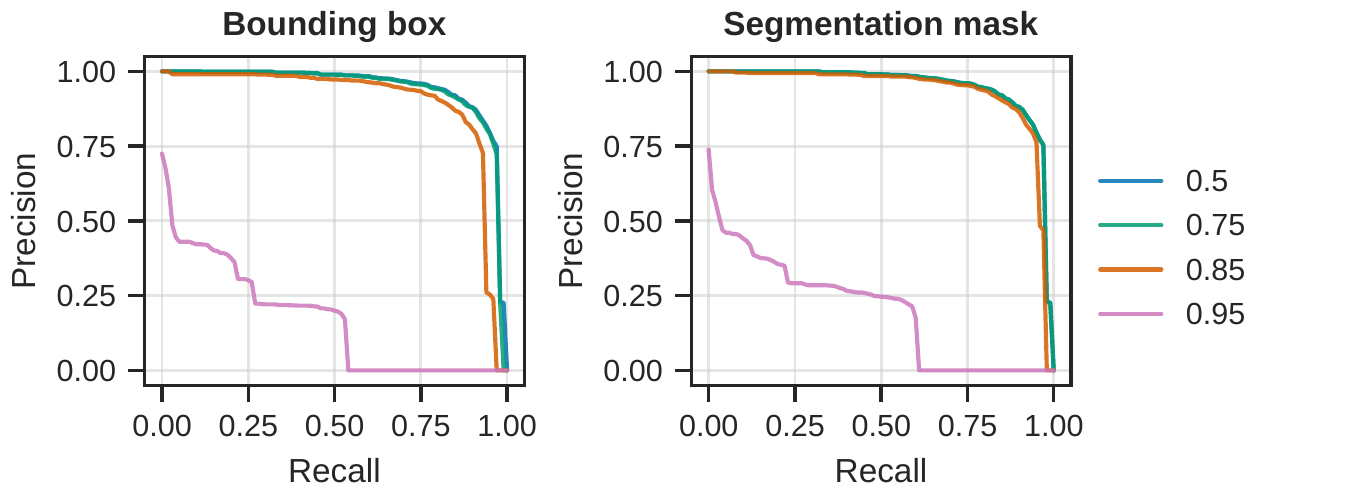}
        \caption{
            The average PR curves without the \enquote{sediment} class, at different IoU thresholds for both tasks.
            \textbf{Left}: Without the \enquote{sediment} class, the curves for thresholds $\qty{0.5, 0.75, 0.85}$ are almost identical,
                whereas there is still a major decrease for the $0.95$ IoU threshold.
            \textbf{Right}: The PR curves for the segmentation task paint the same picture as for the detection task,
                indicating that few predictions are correct above 0.95 IoU, and that very many targets are not being predicted.
        }
        \label{fig:pr_iou_subset}
    \end{figure}

We also wanted to determine how well the model performed at different IoU thresholds, so precision and recall were evaluated for the IoU thresholds $\qty{0.5, 0.75, 0.85, 0.95}$.
Figure~\ref{fig:pr_iou} shows the precision-recall curves for all object classes, and Figure~\ref{fig:pr_iou_subset} shows the sans \enquote{sediment} class curves.
From these results, it is clear that the model performs quite well at IoU thresholds up to and including $0.85$, but at $0.95$ the model does not perform well.

Based on the per-class and per-IoU results, it became evident that some test images containing only \enquote{sediment} class objects, were particularly challenging.
This can in part be explained by the object density in these images, with multiple objects sometimes overlapping or casting shadows on each other.
In the COCO context, these types of object clusters are referred to as a \enquote{crowd}, and receive special treatment during evaluation.
Importantly, none of the objects in our dataset have been annotated as being part of a \enquote{crowd}.\footnote{Due to the resources required to annotate more than 7000 objects based on their proximity to other objects, with sufficient precision and recall.}
Some examples of these dense object clusters can be seen in Figure~\ref{fig:predictions_weird2} and Figure~\ref{fig:predictions_weird3}.
By removing the \enquote{sediment} class from the evaluation, the AP score for the bounding box increased to $0.84$, and for instance segmentation it increased to $0.86$.
Recall also increased significantly, which means that more target objects were correctly detected and segmented.
This increase can be also be seen by comparing the per-IoU curves shown in Figure~\ref{fig:pr_iou_subset} with those in Figure~\ref{fig:pr_iou},
as well as the results presented in Table~\ref{tab:average_scores}.

\subsection{Qualitative Analysis of Predictions}

\begin{figure}[H]
        \includegraphics[width=\linewidth]{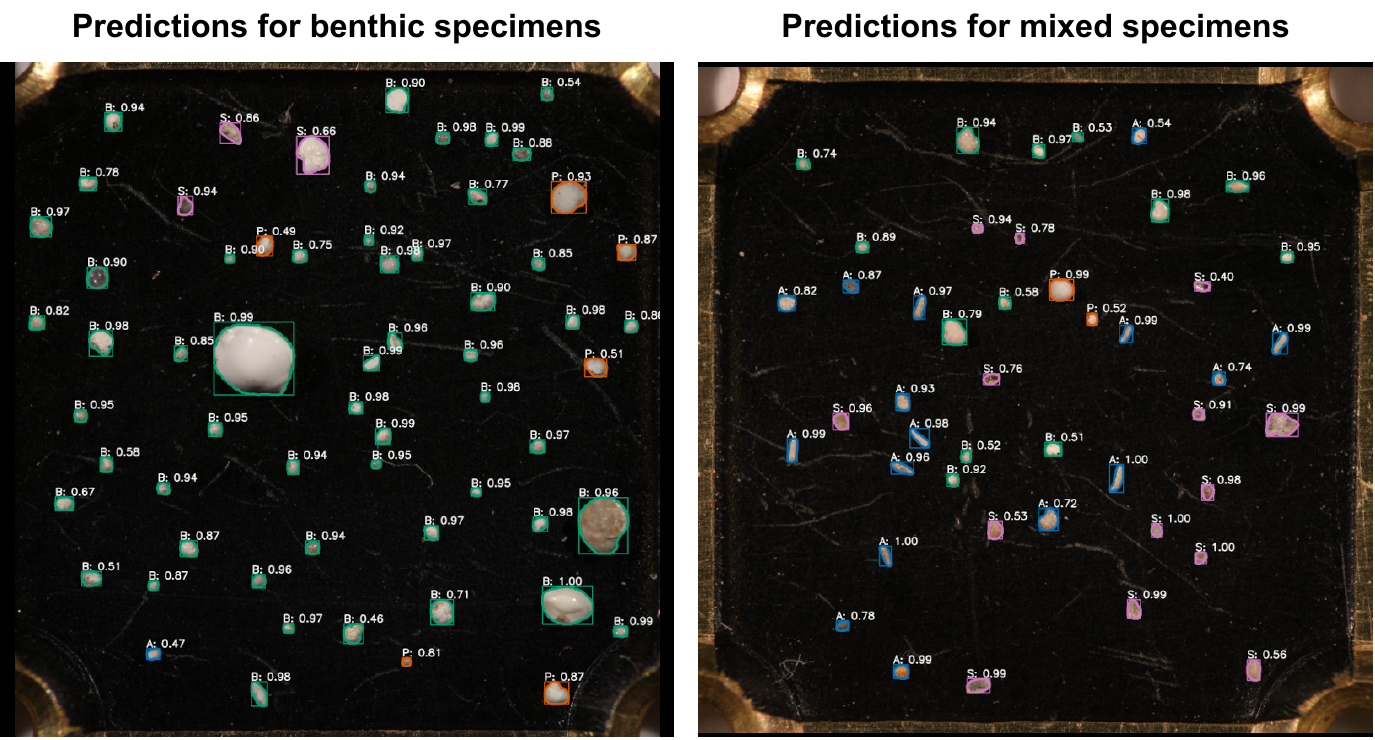}
        \caption{
            Examples of predictions for two images from the test dataset.
            \textbf{Left}: Predictions for purely calcareous benthic specimens. The accuracy and quality of the predicted masks and bounding boxes are good,
                but there are several misclassified objects.
            \textbf{Right}: Predictions for a mixture of specimen types. The accuracy and quality of the predicted masks and bounding boxes are good.
                However, there are misclassified detections for this image as well.
        }
        \label{fig:predictions_1}
    \end{figure}

\begin{figure}[H]
        \includegraphics[width=\linewidth]{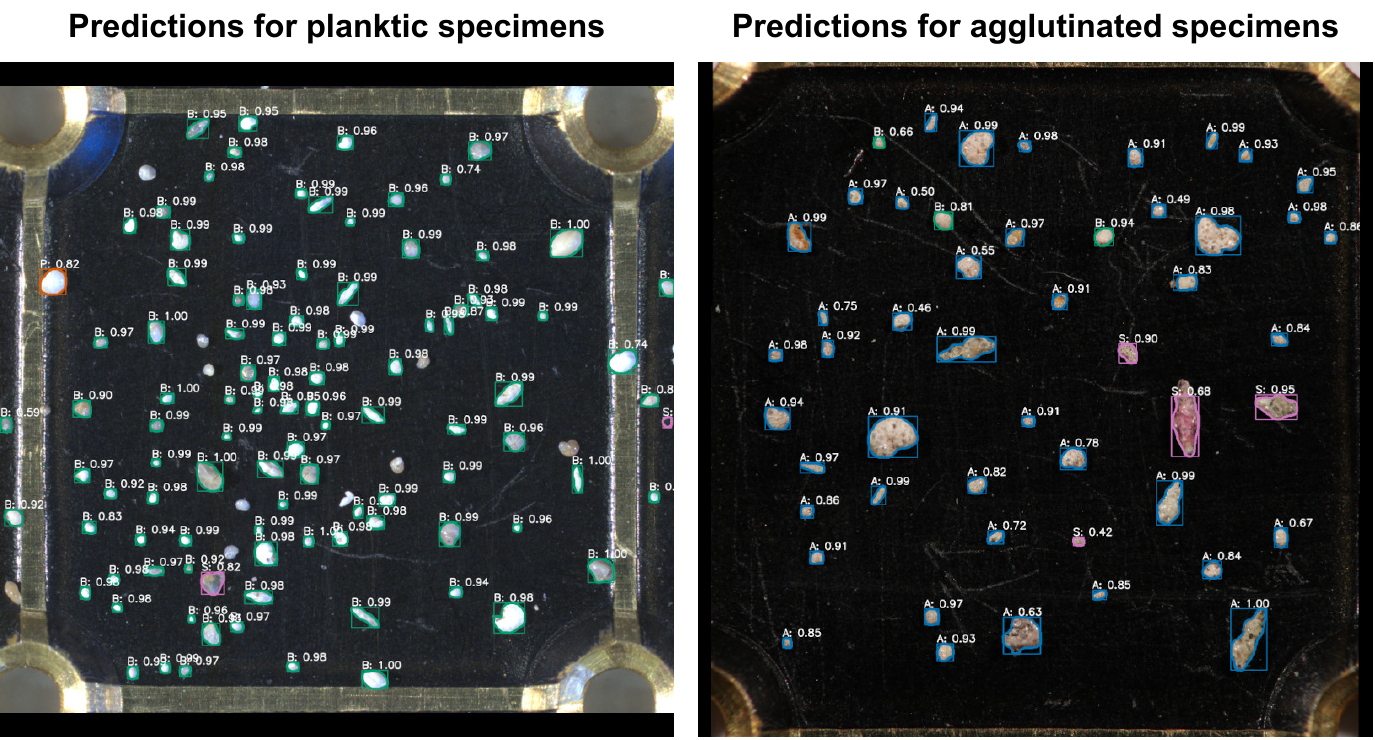}
        \caption{
            Additional examples of predictions for two images from the test dataset.
            \textbf{Left}: Predictions for purely planktic specimens. There are a few false positives, but the accuracy and quality of true positive detections are good.
                Note that some objects that have been misclassified.
            \textbf{Right}: Predictions for purely agglutinated benthic specimens. Good accuracy and quality of predicted masks and bounding boxes the majority of detections.
                However, low quality masks and misclassified detections are visible.
        }
        \label{fig:predictions_2}
    \end{figure}

When evaluating the predictions manually, it became apparent that the overall accuracy and quality of the segmentation masks produced by the model are good.
The boundary of the masks quite precisely delineates the foraminifera and sediment grains from the background.
For the most part, the predicted bounding boxes correspond well with the masks.
One of the biggest challenges seem to lie in the classification of object labels; there are (for trained observers) many obvious misclassifications.
The exact cause is somewhat uncertain, but in many cases the objects are relatively small and feature-less.
It is not hard to image how a feature-less planktic foraminifera can be misclassified as benthic, especially if the object is small.
Other cases of misclassifications are likely caused by a lack of training examples; many seem like out-of-distribution examples due to the high confidence score.
Examples of predictions can be seen in Figure~\ref{fig:predictions_1} and Figure~\ref{fig:predictions_2}.

\section{Discussion} \label{sec:Discussion}

The results presented clearly show that a model built on the Mask R-CNN architecture is capable of performing instance segmentation of microscopic foraminifera.
Using model parameters pretrained on the COCO dataset, we adapted and fine-tuned the model for our novel dataset and achieved AP scores of $0.78$ and $0.80$ on the bounding box and instance segmentation tasks, respectively.
There were significant increases in precision and recall when going from averaging over all IoU thresholds (i.e. AP and AR), to specific IoU thresholds.
When evaluated with an IoU of $0.5$, precision increased to $0.90$ for both tasks, and with an IoU of $0.75$ the precision was $0.88$ for the detection task and $0.90$ for the segmentation task.
This means that predicting bounding boxes and segmentation masks that almost perfectly overlap with their respective ground-truth is challenging for the given dataset, and possibly for the model architecture or hyperparameters.
It is likely that there are errors in the annotated ground-truth object masks in the dataset, both in terms of inaccuracies at the pixel-level, but also potential
false positives or false negatives, meaning that achieving perfect predictions are very unlikely.
Importantly, depending upon the specific application of an instance segmentation model, pixel-perfect predictions might not be a necessity.

Omitting the \enquote{sediment} class also lead to significant increases in model performance, which can be explained by the challenging nature of some test images that contained very dense clusters of sediment grains.
This can in part be mitigated in practical applications by ensuring objects are not clustered, but ideally this also should be addressed at the model-level.
It is possible that this can, to some extent, be overcome by introducing much more training examples with crowded scenes, and correctly annotating all objects as being in a crowd.
Additionally, it is possible that the issue can also be reduced by tuning the hyperparameters of the Mask R-CNN architecture.

Both quantitative and qualitative analysis of the predicted detections and segmentation masks suggest that the model is performing well.
However, the results also show that there are some challenges that should to be investigated further and addressed in future work.

\subsection{Future Research} \label{sec:Future}

Based on the experiments and results, we propose a few research ideas worth investigating in future efforts.

\subsubsection{Expanding and Revisiting the Dataset}

Expanding the dataset is perhaps the most natural extension of the presented work.
If carefully curated, a more exhaustive dataset should help improve some of the corner cases where the model is struggling to produce accurate predictions.
Additionally, with the appropriate resources it would be valuable to ensure every object in the existing dataset is appropriately labeled as part of a \enquote{crowd} or not.
Improving the accuracy of the e.g. densely packed \enquote{sediment} objects, will improve model performance, as well as make the model more applicable to real-world situations.
Another important aspect of expanding the dataset is introduce species-level object classes, as opposed to the high-level categories used today.
Accurately detecting microscopic foraminiferal species is vital to most downstream geoscience applications.

\subsubsection{Additional Hyperparameter Tuning}

If sufficient computational resources are available, performing more exhaustive hyperparameter tuning should be pursued.
While this should include experiments with optimizers, learning rates, and so forth, it should more crucially be focused on the numerous hyperparameters of the Mask R-CNN model components.
Specifically, the parameters of the regional proposal network, and the fully-convolutional network (for mask prediction) should be validated and experimented with.
It is entirely possible some number of these parameters are sub-optimal for the given dataset.

\subsubsection{Improved GPU Training}

While training on multiple GPUs might not lead to big improvements in model performance, the increased effective batch size will help stabilize and speed up training.
Additionally, given the small size of the most objects relative to the image dimensions, training without having to resize the images to fit in GPU memory will increase model performance.
This could be solved directly by using GPUs with more memory, or possibly by partitioning each image across multiple GPUs, predicting on a sub-region per GPU.

\subsubsection{Other Segmentation Models}

We chose to use Mask R-CNN primarily because of its capabilities, but also because proven, pre-trained weights were readily available.
Recently, numerous models have been published that surpass Mask R-CNN in several performance metrics, and importantly also seem to have much faster inference times (which is important for real-world applications.)
Examples of alternative models that should be tested include PANet~\cite{Liu2018}, TensorMask~\cite{Chen2019}, CenterMask~\cite{Wang2020}, SOLOv2~\cite{XWang2020}.

\subsubsection{Uncertainty Estimation}

We have shown that the model is robust to training runs with different random seeds, and the next natural step is to investigate robustness with regards to different training/test data splits,
and to estimate the uncertainty of the model predictions.
Some work has been published on estimating model predictive uncertainty of Mask~R-CNN models~\cite{Miller2017,Miller2019,Morrison2019}.
However, it should be possible to avoid the need for introducing Monte Carlo dropout sampling~\cite{Gal2016}, which requires making changes to existing models, by leveraging the more recent Monte Carlo batch normalization sampling~\cite{Teye2018} technique instead.

\section{Conclusions} \label{sec:Conclusion}

The proposed model achieved an AP of $0.78 \pm 0.00$ on the bounding box (detection) task and $0.80 \pm 0.00$ on the segmentation task, based on 10 training runs with different random seeds.
We also evaluated the model without the challenging sediment grain images, and the AP for both tasks increased to $0.84 \pm 0.00$ and $0.86 \pm 0.00$, respectively.

When evaluating predictions both qualitatively and quantitatively, we saw the predicted bounding boxes and segmentation masks were good for the majority of test cases.
However, there were many cases of incorrect class label predictions; mostly for small objects, or objects that we hypothesize can be considered out-of-distribution.

Based on the presented results, our proposed model for semantic segmentation of microscopic foraminifera, is a step towards automating the process of identifying, counting, and picking of microscopic foraminifera.
However, work remains to be done, such as expanding the dataset to improve the model accuracy, experimenting with other architectures, and implementing uncertainty estimation techniques.

\vspace{6pt}

\authorcontributions{
Conceptualization, T.J., S.S., K.M. and F.G.;
methodology, T.J.;
software, T.J.;
validation, T.J., S.S. and K.M.;
formal analysis, T.J.;
investigation, T.J. and S.S.;
resources, T.J. and S.S.;
data curation, T.J. and S.S.;
writing---original draft preparation, T.J.;
writing---review and editing, T.J., S.S., K.M. and F.G.;
visualization, T.J.;
supervision, K.M. and F.G.;
project administration, T.J.
All authors have read and agreed to the published version of the manuscript.
}

\funding{This research received no external funding. The APC was funded by UiT The Arctic University of Norway.}


\conflictsofinterest{The authors declare no conflict of interest.} 

\appendixtitles{yes}
\appendixstart
\appendix

\appendix
\clearpage
\section{Prediction Examples}


    \end{paracol}
    \begin{paracol}{1}
    \appendix
    \begin{figure}[H]
        \widefigure
        \includegraphics[width=\linewidth]{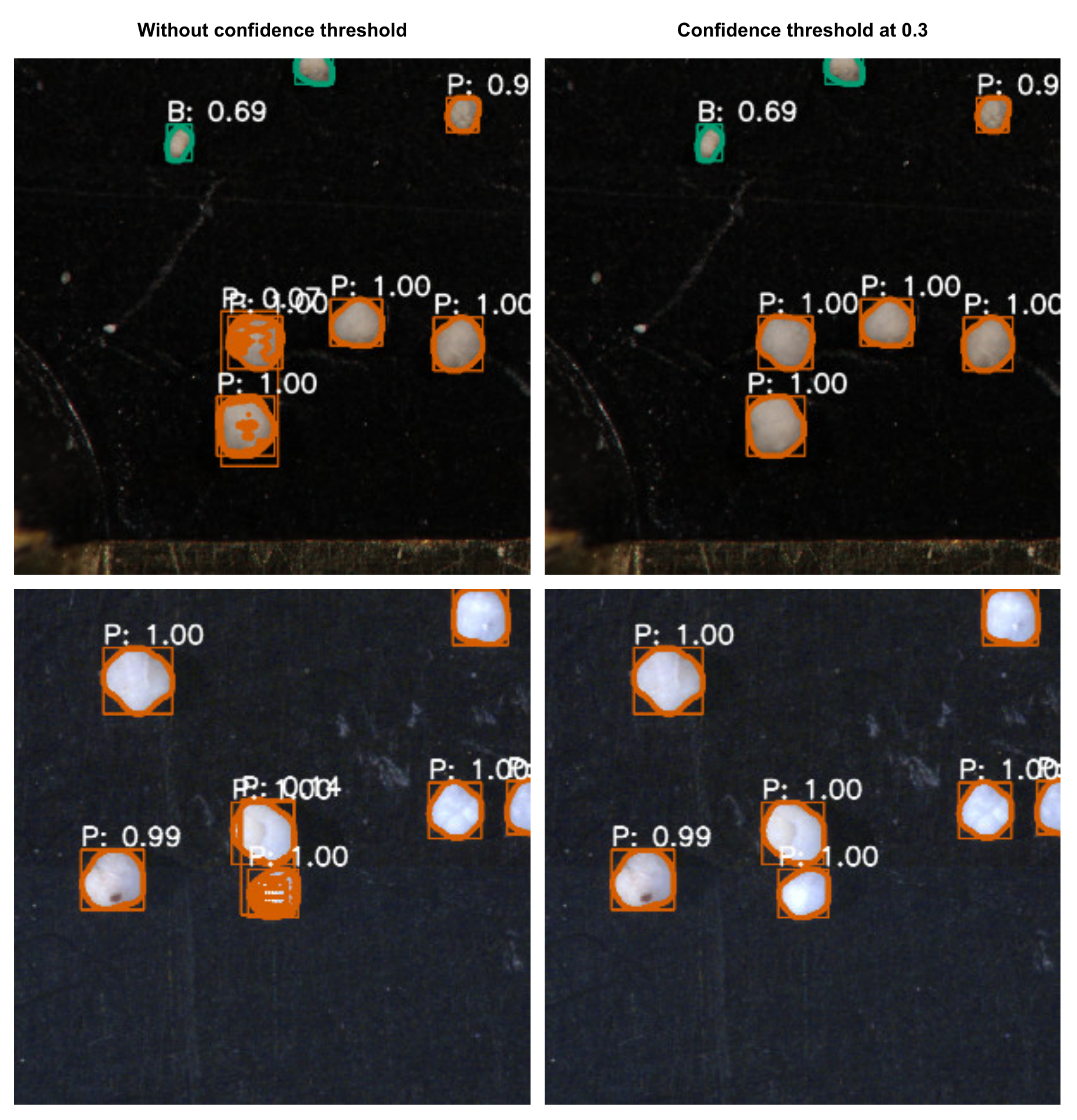}
        \caption{
            Examples of predicted bounding boxes and segmentation masks for the \enquote{planktic} class.
            \textbf{Left column}: Overlapping predictions can be seen near the middle of both images.
                The confidence score for the overlapping predictions with low-quality masks is significantly lower than the high-quality predictions.
                We can also see that some smaller objects in the top image have been misclassified as the \enquote{benthic} class.
            \textbf{Right column}: The overlapping predictions have been removed by thresholding the confidence score at $0.3$.
        }
        \label{fig:predictions_weird1}
    \end{figure}
    
    \begin{figure}[H]
        \widefigure
        \includegraphics[width=\linewidth]{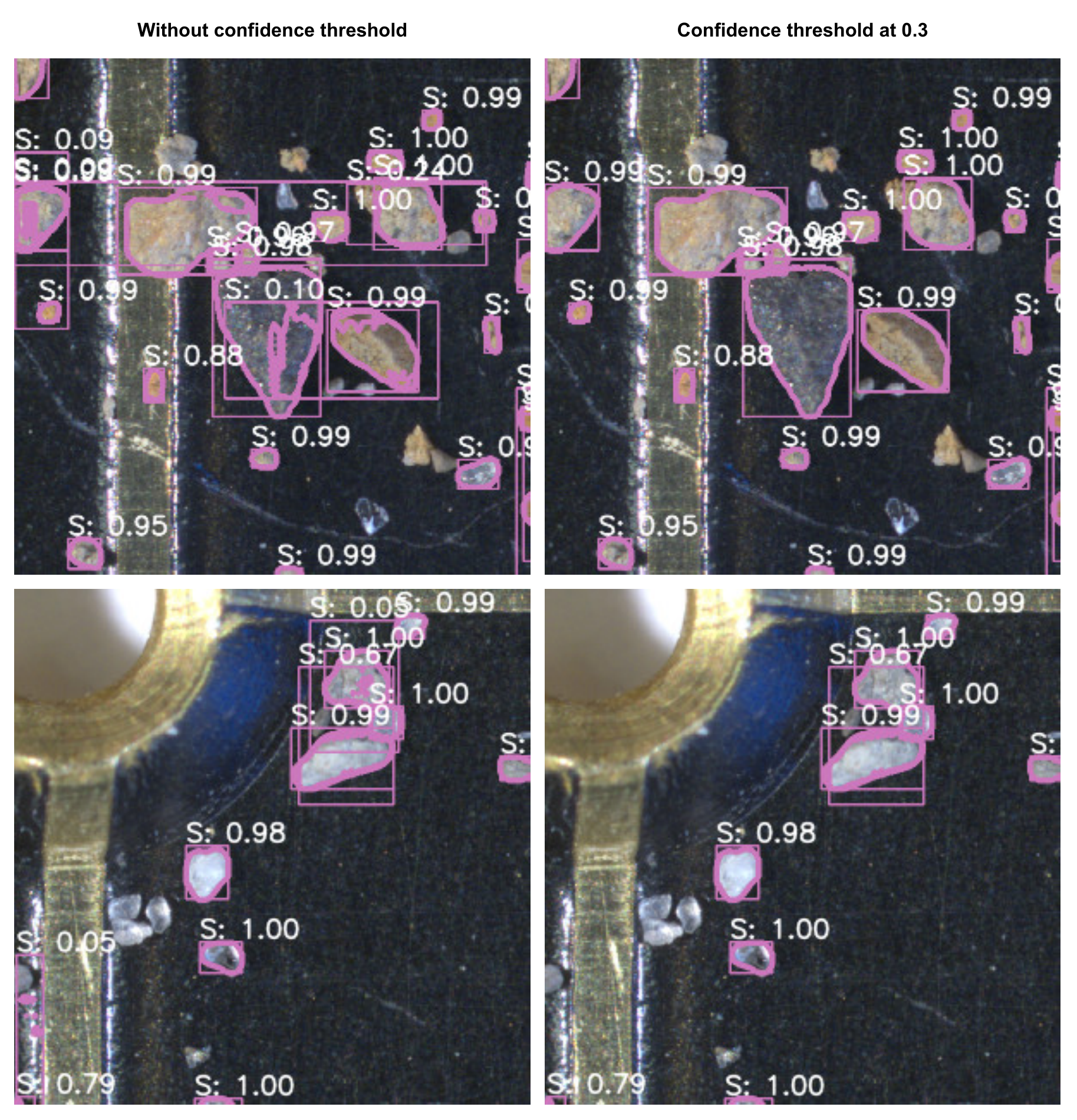}
        \caption{
            Examples of predicted bounding boxes and segmentation masks for the \enquote{sediment} class.
            \textbf{Left column}: Overlapping predictions can be seen near the middle of both images.
                Also, notice that several objects have been missed entirely.
            \textbf{Right column}: Most of the overlapping predictions have been removed by thresholding the confidence score at $0.3$.
        }
        \label{fig:predictions_weird2}
    \end{figure}
    
    \begin{figure}[H]
        \widefigure
        \includegraphics[width=\linewidth]{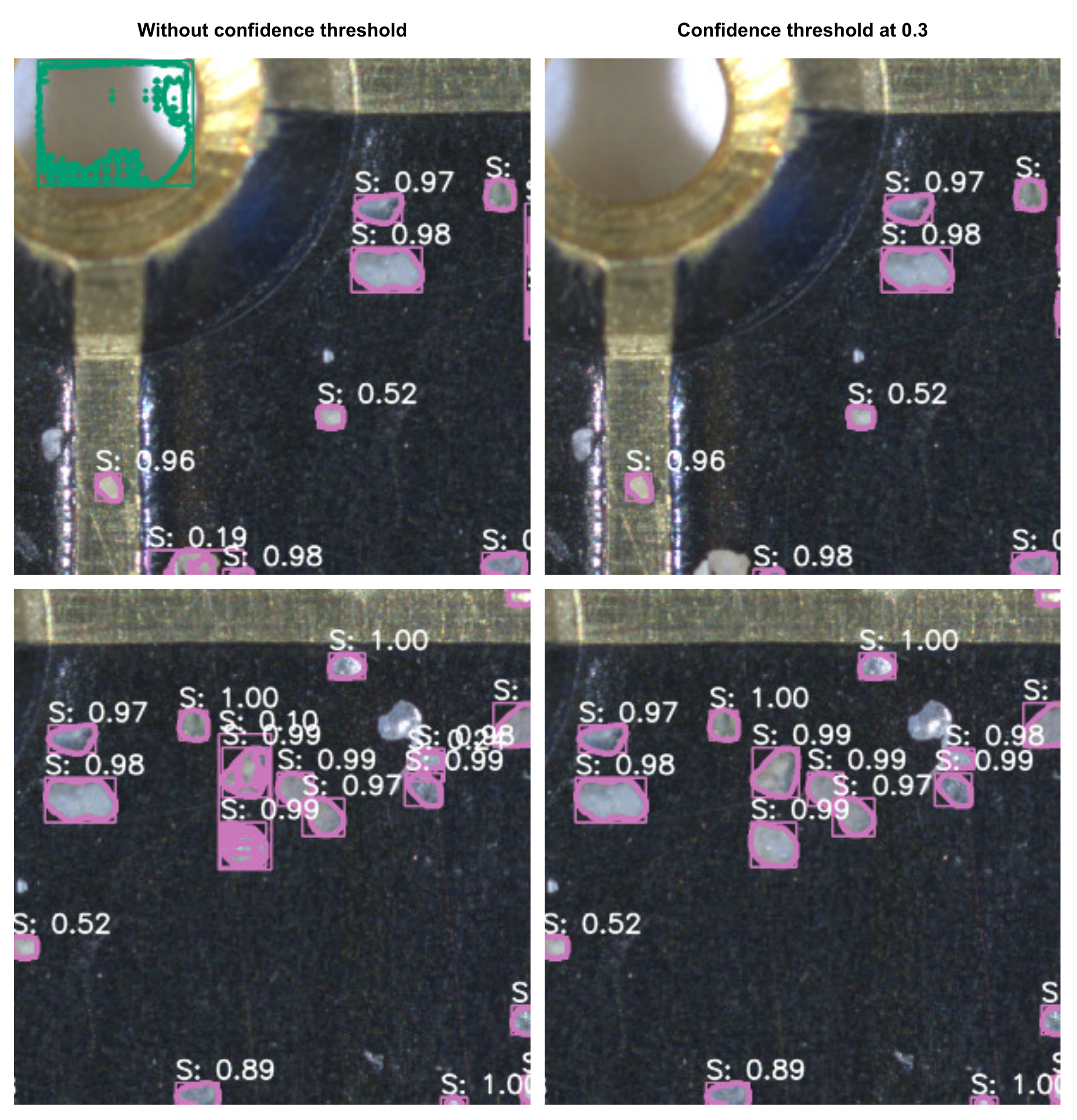}
        \caption{
            Additional examples of predicted bounding boxes and segmentation masks for the \enquote{sediment} class.
            \textbf{Left column}: A very obvious false positive detection of the \enquote{benthic} class can be seen near the top-left corner of the first image.
                For the second image, several overlapping predictions can be seen.
            \textbf{Right column}: The false positive detection and the overlapping predictions have been removed by thresholding the confidence score at $0.3$.
        }
        \label{fig:predictions_weird3}
    \end{figure}
    \end{paracol}
    \begin{paracol}{2}
    \switchcolumn
    \ifcsstring{@status}{submit}{\linenumbers}{}

\end{paracol}
\clearpage
\reftitle{References}


\externalbibliography{yes}
\bibliography{refs.bib}

\begin{thebibliography}{999}

\bibitem[{Hayward, Bruce, B.W.} \em{et~al.}(2021){Hayward, Bruce, B.W.}, {Le
  Coze, Fran\c{c}ois, F.}, {Vachard, Daniel, D.}, and {Gross, Onno,
  O.}]{Hayward2021}
{Hayward, Bruce, B.W.}.; {Le Coze, Fran\c{c}ois, F.}.; {Vachard, Daniel, D.}.;
  {Gross, Onno, O.}.
\newblock World Foraminifera Database. Accessed at
  http://www.marinespecies.org/foraminifera on 2021-04-08,  2021.
\newblock
  doi:{\changeurlcolor{black}\href{https://doi.org/10.14284/305}{\detokenize{10.14284/305}}}.

\bibitem[Emiliani(1955)]{Emiliani1955}
Emiliani, C.
\newblock Pleistocene temperatures.
\newblock {\em The Journal of Geology} {\bf 1955}, {\em 63},~538--578.

\bibitem[Hald \em{et~al.}(2007)Hald, Andersson, Ebbesen, Jansen,
  Klitgaard-Kristensen, Risebrobakken, Salomonsen, Sarnthein, Sejrup, and
  Telford]{Hald2007}
Hald, M.; Andersson, C.; Ebbesen, H.; Jansen, E.; Klitgaard-Kristensen, D.;
  Risebrobakken, B.; Salomonsen, G.R.; Sarnthein, M.; Sejrup, H.P.; Telford,
  R.J.
\newblock Variations in temperature and extent of Atlantic Water in the
  northern North Atlantic during the Holocene.
\newblock {\em Quaternary Science Reviews} {\bf 2007}, {\em 26},~3423--3440.

\bibitem[Katz \em{et~al.}(2010)Katz, Cramer, Franzese, Hönisch, Miller,
  Rosenthal, and Wright]{Katz2010}
Katz, M.E.; Cramer, B.S.; Franzese, A.; Hönisch, B.; Miller, K.G.; Rosenthal,
  Y.; Wright, J.D.
\newblock Traditional and emerging geochemical proxies in foraminifera.
\newblock {\em The Journal of Foraminiferal Research} {\bf 2010}, {\em
  40},~165--192.

\bibitem[Suokhrie \em{et~al.}(2017)Suokhrie, Saraswat, Nigam, Kathal, and
  Talib]{Suokhrie2017}
Suokhrie, T.; Saraswat, R.; Nigam, R.; Kathal, P.; Talib, A.
\newblock Foraminifera as bio-indicators of pollution: a review of research
  over the last decade.
\newblock {\em Micropaleontology and its Applications. Scientific Publishers
  (India)} {\bf 2017}, pp. 265--284.

\bibitem[Zhong \em{et~al.}(2017)Zhong, Ge, Kanakiya, Marchitto, and
  Lobaton]{Zhong2017}
Zhong, B.; Ge, Q.; Kanakiya, B.; Marchitto, R.M.T.; Lobaton, E.
\newblock A comparative study of image classification algorithms for
  Foraminifera identification.
\newblock  2017 IEEE Symposium Series on Computational Intelligence (SSCI).
  IEEE,  2017, pp. 1--8.

\bibitem[Ge \em{et~al.}(2017)Ge, Zhong, Kanakiya, Mitra, Marchitto, and
  Lobaton]{Ge2017}
Ge, Q.; Zhong, B.; Kanakiya, B.; Mitra, R.; Marchitto, T.; Lobaton, E.
\newblock Coarse-to-fine foraminifera image segmentation through 3D and deep
  features.
\newblock  2017 IEEE Symposium Series on Computational Intelligence (SSCI).
  IEEE,  2017, pp. 1--8.

\bibitem[de~Garidel-Thoron \em{et~al.}(2017)de~Garidel-Thoron, Marchant, Soto,
  Gally, Beaufort, Bolton, Bouslama, Licari, Mazur, Brutti, et~al.]{De2017}
de~Garidel-Thoron, T.; Marchant, R.; Soto, E.; Gally, Y.; Beaufort, L.; Bolton,
  C.T.; Bouslama, M.; Licari, L.; Mazur, J.C.; Brutti, J.M.; others.
\newblock Automatic Picking of Foraminifera: Design of the Foraminifera Image
  Recognition and Sorting Tool (FIRST) Prototype and Results of the Image
  Classification Scheme.
\newblock  AGU Fall Meeting Abstracts,  2017.

\bibitem[Mitra \em{et~al.}(2019)Mitra, Marchitto, Ge, Zhong, Kanakiya, Cook,
  Fehrenbacher, Ortiz, Tripati, and Lobaton]{Mitra2019}
Mitra, R.; Marchitto, T.; Ge, Q.; Zhong, B.; Kanakiya, B.; Cook, M.;
  Fehrenbacher, J.; Ortiz, J.; Tripati, A.; Lobaton, E.
\newblock Automated species-level identification of planktic foraminifera using
  convolutional neural networks, with comparison to human performance.
\newblock {\em Marine Micropaleontology} {\bf 2019}, {\em 147},~16--24.

\bibitem[Johansen and Sørensen(2020)]{Johansen2020}
Johansen, T.H.; Sørensen, S.A.
\newblock Towards detection and classification of microscopic foraminifera
  using transfer learning.
\newblock  Proceedings of the Northern Lights Deep Learning Workshop,  2020.

\bibitem[He \em{et~al.}(2017)He, Gkioxari, Doll{\'{a}}r, and Girshick]{He2017}
He, K.; Gkioxari, G.; Doll{\'{a}}r, P.; Girshick, R.
\newblock {Mask R-CNN}.
\newblock {\em IEEE Transactions on Pattern Analysis and Machine Intelligence}
  {\bf 2017}, {\em 42},~386--397,
  \href{http://xxx.lanl.gov/abs/1703.06870}{{\normalfont [1703.06870]}}.
\newblock
  doi:{\changeurlcolor{black}\href{https://doi.org/10.1109/TPAMI.2018.2844175}{\detokenize{10.1109/TPAMI.2018.2844175}}}.

\bibitem[Dutta \em{et~al.}(2016)Dutta, Gupta, and Zissermann]{Dutta2016}
Dutta, A.; Gupta, A.; Zissermann, A.
\newblock {VGG} Image Annotator ({VIA}).
\newblock http://www.robots.ox.ac.uk/~vgg/software/via/,  2016.
\newblock Version: 2.0.9, Accessed: 2020-03-10.

\bibitem[Dutta and Zisserman(2019)]{Dutta2019}
Dutta, A.; Zisserman, A.
\newblock The {VIA} Annotation Software for Images, Audio and Video.
\newblock  Proceedings of the 27th ACM International Conference on Multimedia;
  ACM: New York, NY, USA,  2019; MM '19.
\newblock
  doi:{\changeurlcolor{black}\href{https://doi.org/10.1145/3343031.3350535}{\detokenize{10.1145/3343031.3350535}}}.

\bibitem[Girshick(2015)]{Girshick2015}
Girshick, R.
\newblock {Fast R-CNN}.
\newblock  2015 IEEE International Conference on Computer Vision (ICCV). IEEE,
  2015, pp. 1440--1448,  \href{http://xxx.lanl.gov/abs/1504.08083}{{\normalfont
  [1504.08083]}}.
\newblock
  doi:{\changeurlcolor{black}\href{https://doi.org/10.1109/ICCV.2015.169}{\detokenize{10.1109/ICCV.2015.169}}}.

\bibitem[Ren \em{et~al.}(2017)Ren, He, Girshick, and Sun]{Ren2017}
Ren, S.; He, K.; Girshick, R.; Sun, J.
\newblock {Faster R-CNN: Towards Real-Time Object Detection with Region
  Proposal Networks}.
\newblock {\em IEEE Transactions on Pattern Analysis and Machine Intelligence}
  {\bf 2017}, {\em 39},~1137--1149,
  \href{http://xxx.lanl.gov/abs/1506.01497}{{\normalfont [1506.01497]}}.
\newblock
  doi:{\changeurlcolor{black}\href{https://doi.org/10.1109/TPAMI.2016.2577031}{\detokenize{10.1109/TPAMI.2016.2577031}}}.

\bibitem[Liu \em{et~al.}(2018)Liu, Qi, Qin, Shi, and Jia]{Liu2018}
Liu, S.; Qi, L.; Qin, H.; Shi, J.; Jia, J.
\newblock {Path Aggregation Network for Instance Segmentation}.
\newblock  2018 IEEE/CVF Conference on Computer Vision and Pattern Recognition.
  IEEE,  2018, pp. 8759--8768,
  \href{http://xxx.lanl.gov/abs/1803.01534}{{\normalfont [1803.01534]}}.
\newblock
  doi:{\changeurlcolor{black}\href{https://doi.org/10.1109/CVPR.2018.00913}{\detokenize{10.1109/CVPR.2018.00913}}}.

\bibitem[Chen \em{et~al.}(2019)Chen, Girshick, He, and Dollar]{Chen2019}
Chen, X.; Girshick, R.; He, K.; Dollar, P.
\newblock {TensorMask: A Foundation for Dense Object Segmentation}.
\newblock  2019 IEEE/CVF International Conference on Computer Vision (ICCV).
  IEEE,  2019, Vol. 2019-Octob, pp. 2061--2069,
  \href{http://xxx.lanl.gov/abs/1903.12174}{{\normalfont [1903.12174]}}.
\newblock
  doi:{\changeurlcolor{black}\href{https://doi.org/10.1109/ICCV.2019.00215}{\detokenize{10.1109/ICCV.2019.00215}}}.

\bibitem[Wang \em{et~al.}(2020{\natexlab{a}})Wang, Xu, Shen, Cheng, and
  Yang]{Wang2020}
Wang, Y.; Xu, Z.; Shen, H.; Cheng, B.; Yang, L.
\newblock {CenterMask: single shot instance segmentation with point
  representation}.
\newblock {\em Proceedings of the IEEE Computer Society Conference on Computer
  Vision and Pattern Recognition} {\bf 2020}, pp. 12190--12199,
  \href{http://xxx.lanl.gov/abs/2004.04446}{{\normalfont [2004.04446]}}.

\bibitem[Wang \em{et~al.}(2020{\natexlab{b}})Wang, Zhang, Kong, Li, and
  Shen]{XWang2020}
Wang, X.; Zhang, R.; Kong, T.; Li, L.; Shen, C.
\newblock {SOLOv2: Dynamic and Fast Instance Segmentation}.
\newblock  NeurIPS,  2020, pp. 1--17,
  \href{http://xxx.lanl.gov/abs/2003.10152}{{\normalfont [2003.10152]}}.

\bibitem[He \em{et~al.}(2015)He, Zhang, Ren, and Sun]{He2015}
He, K.; Zhang, X.; Ren, S.; Sun, J.
\newblock {Deep Residual Learning for Image Recognition}.
\newblock {\em Proceedings of the IEEE Computer Society Conference on Computer
  Vision and Pattern Recognition} {\bf 2015}, {\em 2016-Decem},~770--778,
  \href{http://xxx.lanl.gov/abs/1512.03385}{{\normalfont [1512.03385]}}.
\newblock
  doi:{\changeurlcolor{black}\href{https://doi.org/10.1109/CVPR.2016.90}{\detokenize{10.1109/CVPR.2016.90}}}.

\bibitem[Lin \em{et~al.}(2017)Lin, Dollar, Girshick, He, Hariharan, and
  Belongie]{Lin2017}
Lin, T.Y.; Dollar, P.; Girshick, R.; He, K.; Hariharan, B.; Belongie, S.
\newblock {Feature Pyramid Networks for Object Detection}.
\newblock  2017 IEEE Conference on Computer Vision and Pattern Recognition
  (CVPR). IEEE,  2017, pp. 936--944,
  \href{http://xxx.lanl.gov/abs/1612.03144}{{\normalfont [1612.03144]}}.
\newblock
  doi:{\changeurlcolor{black}\href{https://doi.org/10.1109/CVPR.2017.106}{\detokenize{10.1109/CVPR.2017.106}}}.

\bibitem[Lin \em{et~al.}(2015)Lin, Maire, Belongie, Bourdev, Girshick, Hays,
  Perona, Ramanan, Zitnick, and Dollár]{Lin2015}
Lin, T.Y.; Maire, M.; Belongie, S.; Bourdev, L.; Girshick, R.; Hays, J.;
  Perona, P.; Ramanan, D.; Zitnick, C.L.; Dollár, P.
\newblock {Microsoft COCO: Common Objects in Context},  2015,
  \href{http://xxx.lanl.gov/abs/1405.0312}{{\normalfont
  [arXiv:cs.CV/1405.0312]}}.

\bibitem[Sutskever \em{et~al.}(2013)Sutskever, Martens, Dahl, and
  Hinton]{Sutskever2013}
Sutskever, I.; Martens, J.; Dahl, G.; Hinton, G.
\newblock {On the importance of initialization and momentum in deep learning}.
\newblock  Proceedings of the 30th International Conference on Machine
  Learning; McAllester, S.D.; David., Eds. PMLR,  2013, Vol.~28, {\em
  Proceedings of Machine Learning Research}, pp. 1139--1147.

\bibitem[Kingma and Ba(2015)]{Kingma2015}
Kingma, D.P.; Ba, J.
\newblock {Adam: A Method for Stochastic Optimization}.
\newblock  International Conference on Learning Representations,  2015,
  \href{http://xxx.lanl.gov/abs/1412.6980}{{\normalfont [1412.6980]}}.

\bibitem[Loshchilov and Hutter(2019)]{Loshchilov2019}
Loshchilov, I.; Hutter, F.
\newblock {Decoupled Weight Decay Regularization}.
\newblock  International Conference on Learning Representations,  2019,
  \href{http://xxx.lanl.gov/abs/1711.05101}{{\normalfont [1711.05101]}}.

\bibitem[COC()]{COCO2021}
{COCO: Common Object in Context -- Detection Evaluation}.
\newblock \url{https://cocodataset.org/#detection-eval}.
\newblock Accessed: 2021-04-17.

\bibitem[Miller \em{et~al.}(2018)Miller, Nicholson, Dayoub, and
  Sünderhauf]{Miller2017}
Miller, D.; Nicholson, L.; Dayoub, F.; Sünderhauf, N.
\newblock Dropout Sampling for Robust Object Detection in Open-Set Conditions.
\newblock  2018 IEEE International Conference on Robotics and Automation
  (ICRA),  2018, pp. 3243--3249.
\newblock
  doi:{\changeurlcolor{black}\href{https://doi.org/10.1109/ICRA.2018.8460700}{\detokenize{10.1109/ICRA.2018.8460700}}}.

\bibitem[Miller \em{et~al.}(2019)Miller, Dayoub, Milford, and
  Sünderhauf]{Miller2019}
Miller, D.; Dayoub, F.; Milford, M.; Sünderhauf, N.
\newblock Evaluating Merging Strategies for Sampling-based Uncertainty
  Techniques in Object Detection.
\newblock  2019 International Conference on Robotics and Automation (ICRA),
  2019, pp. 2348--2354.
\newblock
  doi:{\changeurlcolor{black}\href{https://doi.org/10.1109/ICRA.2019.8793821}{\detokenize{10.1109/ICRA.2019.8793821}}}.

\bibitem[Morrison \em{et~al.}(2019)Morrison, Milan, and
  Antonakos]{Morrison2019}
Morrison, D.; Milan, A.; Antonakos, N.
\newblock {Uncertainty-aware Instance Segmentation using Dropout Sampling}.
\newblock  The Robotic Vision Probabilistic Object Detection Challenge (CVPR
  2019 Workshop),  2019.

\bibitem[Gal and Ghahramani(2016)]{Gal2016}
Gal, Y.; Ghahramani, Z.
\newblock Dropout as a Bayesian Approximation: Representing Model Uncertainty
  in Deep Learning.
\newblock  Proceeding of the 33rd International Conference on Machine Learning.
  PMLR,  2016, Vol.~48, pp. 1050--1059.

\bibitem[Teye \em{et~al.}(2018)Teye, Azizpour, and Smith]{Teye2018}
Teye, M.; Azizpour, H.; Smith, K.
\newblock {Bayesian Uncertainty Estimation for Batch Normalized Deep Networks}.
\newblock  Proc. 35th Int. Conf. Mach. Learn.,  2018, Vol.~11, pp. 7824--7833,
  \href{http://xxx.lanl.gov/abs/1802.06455}{{\normalfont [1802.06455]}}.

\end{thebibliography}

\end{document}